\documentclass[letterpaper, 10 pt, conference]{ieeeconf}  

\IEEEoverridecommandlockouts                              

\usepackage{multicol}
\usepackage{graphicx}
\usepackage[font=small,labelfont=bf,justification=justified]{caption}
\usepackage{subcaption}
\usepackage{amsmath}
\usepackage{amssymb}
\usepackage{textcomp}
\usepackage{gensymb}
\usepackage{bm}
\usepackage{float}
\usepackage{color}
\usepackage{multirow}
\usepackage{hhline}
\usepackage{changepage}
\usepackage{comment}
\usepackage{titlesec}
\titlespacing\section{0pt}{3pt}{3pt}

\usepackage{cite}

\usepackage[ruled,vlined,linesnumbered]{algorithm2e}

\usepackage[bookmarks=true]{hyperref}

\newcommand{\nonl}{\renewcommand{\nl}{\let\nl\oldnl}}

\title{\LARGE \bf
Containerized Vertical Farming Using Cobots
}
\author{Dasharadhan Mahalingam$^1$, Aditya Patankar$^1$, Khiem Phi$^2$, Nilanjan Chakraborty$^1$,  \\ Ryan McGann$^3$, and IV Ramakrishnan$^2$
\date{}
\thanks{$^{1}$The authors are with the Department of Mechanical Engineering, 
        Stony Brook University, USA.
       {\tt\small \{aditya.patankar, dasharadhan.mahalingam, nilanjan.chakraborty\}@stonybrook.edu.}}%
\thanks{$^{2}$The authors are with the Department of Computer Science, 
        Stony Brook University, USA.
       {\tt\small \{kphi, ram\}@cs.stonybrook.edu.}}%
\thanks{$^{3}$The author is with CubicAcres LLC, 
        Stony Brook, USA.
       {\tt\small ryan@cubic-acres.com}}%
}

\begin{document}

\maketitle

\begin{abstract}
Containerized vertical farming is a type of vertical farming practice using hydroponics in which plants are grown in vertical layers within a mobile shipping container. Space limitations within shipping containers make the automation of different farming operations challenging. In this paper, we explore the use of cobots (i.e., collaborative robots) to automate two key farming operations, namely, the transplantation of saplings and the harvesting of grown plants. Our method uses a single demonstration from a farmer to extract the motion constraints associated with the tasks, namely, transplanting and harvesting, and can then generalize to different instances of the same task. For transplantation, the motion constraint arises during insertion of the sapling within the growing tube, whereas for harvesting, it arises during extraction from the growing tube. We present experimental results to show that using RGBD camera images (obtained from an eye-in-hand configuration) and one demonstration for each task, it is feasible to perform transplantation of saplings and harvesting of leafy greens using a cobot, without task-specific programming.

\textit{Video}--- \url{https://youtu.be/KMqA-4GvKwk}
\end{abstract}

\section{Introduction}
\begin{figure}[t!]
  \centering
  \includegraphics[width=\linewidth]{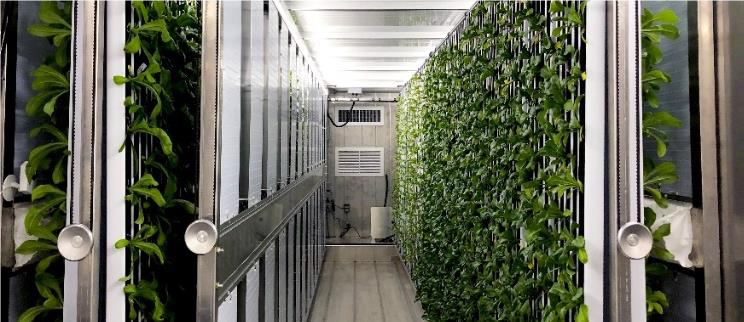}
  \caption{Vertical Farm in a shipping container with vertical grow panels stacked horizontally.}
 \label{fig:vertical_farm} 
\end{figure}

Vertical hydroponic farming is a farming practice in which crops, typically leafy greens, are grown in an indoor environment with controlled temperature and lighting. There are two ways in which the crops are grown: (a) in horizontal layers that are stacked vertically (usually used in large abandoned warehouses and buildings), and (b) in vertical layers that are stacked horizontally (usually used in mobile shipping containers; see Figure~\ref{fig:vertical_farm}). We will call the latter Containerized Vertical Farming (CVF). 
Apart from sharing other environmental benefits of vertical farming~\cite{birkby2016vertical, marchant2017robotic}, CVF ensures hyperlocal food production as these mobile containers can be located directly at the point of consumption, thereby significantly reducing transportation overheads and ensuring longer shelf life and freshness of produce. 

However, current CVF practice is labor intensive, especially because it requires key agricultural steps such as transplantation of saplings, harvesting, and crop inspection to be performed manually. In CVF, space constraints of the shipping containers, as well as the different layout of plants, preclude the use of automated devices~\cite{plenty_main, plenty_tigris_1, bowery_main, bowery_bloomberg, bowery_tech_and_innovation} built for vertical farming applications in warehouses (or large buildings). Therefore, {\em the objective of this paper is to explore the use of commercial standard cobots (i.e., collaborative robots) to automate key vertical farming operations in CVF}.

\begin{figure}[t!]
  \centering
  \includegraphics[width=\linewidth]{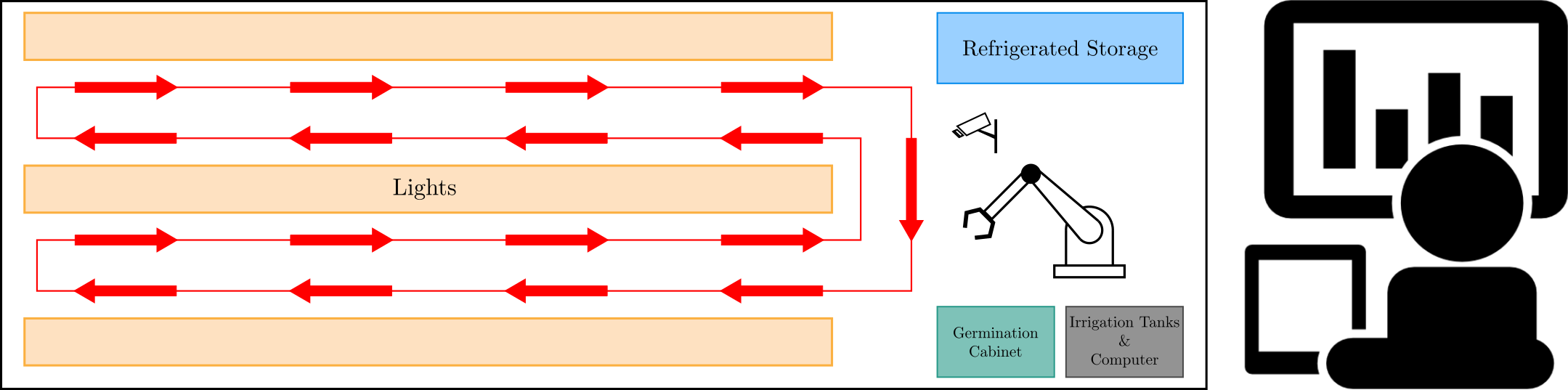}
  \caption{Schematic sketch of our CVF with grow panels moving on conveyors, cobot workstation, and remotely monitoring farmer.}
 \label{fig:layout} 
\end{figure}

Figure~\ref{fig:vertical_farm} shows a typical CVF where the spacing between the grow panels is so small that the panels must be manually moved apart for a human to go in (Figure~\ref{fig:vertical_farm} shows such a configuration in which the two middle panels have been moved to make enough space for a human to move in). Apart from precluding the use of automated devices recently developed for vertical farming in large warehouses, this setup with tight space constraints also precludes the use of mobile manipulators (possibly on rails) that are traditionally used in harvesting operations in open farms or greenhouses. Figure~\ref{fig:layout} shows a schematic sketch of our envisioned CVF, where the vertical grow panels move on conveyors and stop at a robotic workstation where the robot performs the farming tasks.

Conveyors that can carry the grow panels exist today. However, one key challenge in realizing our vision of CVF is for the robot to perform reliably the various manipulation tasks involved in farming using (RGBD) camera images. For this paper, we consider two manipulation tasks, namely, \textit{\textbf{transplanting of saplings}} and \textit{\textbf{harvesting}} (by removing the whole plant with the root intact).  Both of these tasks require constrained motion of the robot end effector, which is a challenging problem, especially when estimation of the constraints have to be based on the image data. For transplantation, the motion constraint arises during insertion of the sapling within the growing tube, whereas for harvesting, it arises during extraction from the growing tube. In both cases, the explicit representation of the constraints depends on the pose of the slots in the growing tube, which have to be estimated from images. Thus, programming the constraints required to execute the task successfully requires someone with specialized knowledge in robotics, which most farmers will not have. Therefore, it is hard for a farmer to re-purpose general purpose manipulators for performing tasks in a CVF. 

However, for cobots, a farmer can easily hold the hand of the robot and show it how to perform a task (in zero-gravity mode). A single successful demonstration implicitly contains the constraints the end-effector motion should satisfy during task execution. Therefore, in principle, it is possible to use this single demonstration to plan a path for the robot to perform different instances of the task (e.g., use a demonstration of inserting a sapling in one slot to insert it in another slot). We present a novel method for manipulation planning for transplantation and harvesting tasks that combines (a) a deep learning-based foundation model for image segmentation, namely, the Segment Anything Model (SAM) from Meta AI~\cite{kirillov2023segmentanything} (b) geometric knowledge of the slots in the growing tubes, and (c) a screw-geometric representation of the demonstration as a sequence of constant screw motions or one-parameter motion subgroups of $SE(3)$, the group of rigid body motions, as proposed in~\cite{mahalingam2023humanguided}. Using SAM, the geometry of the slot openings, and the kinesthetic demonstration, we obtain the constraints for the new task instance as motion subgroup constraints, which can be satisfied using Screw Linear Interpolation (ScLERP) as a basic motion planner~\cite{sclerp_motion_planner}. We present experimental results to show that our method can indeed generalize from one demonstration to perform transplantation and harvesting tasks.


\section{Related Work}

Robotic and AI-based technologies have been developed and used extensively
for open-field farming~\cite{bac2014harvesting, roldan2018robots, vasconez2019human}. These technologies range from
automated monitoring and inspection~\cite{van2017multisensor, corollaro2014combined, pace2011relationship, munera2017astringency, mishraIoT2022, ukwuoma2022recent, tang2020recognition}, harvesting~\cite{nuske2011visual, nuske2011yield, de2011design, bac2014harvesting, ukwuoma2022recent} to post-harvest
processing of the produce~\cite{corollaro2014combined, pace2011relationship, ukwuoma2022recent, tang2020recognition}.
Robotic technologies have also been developed and deployed in greenhouses~\cite{bagagiolo2022greenhouse} specifically for the purposes of fruit and crop harvesting~\cite{van2013robotics,sa2017peduncle,vitzrabin2016changing} and weeding~\cite{aastrand2002agricultural ,slaughter2008autonomous, heravi2018development}.
A key point to note is that the existing robotic technologies developed for open-field farming or greenhouses cannot be used directly for automation in Containerized Vertical Farms (CVFs) mainly due to the drastically different layout and design of cultivation systems along with space constraints. 

Recent work on introducing automation~\cite{warrier2021design}, robotics~\cite{marchant2017robotic, lauguico2019implementation, chitre2023optimal} or AI-based technologies for vertical farming has focused mainly on the development of live monitoring and inspection systems \cite{sreedevi2020digital, samaranayake2022autonomous, gubanov2022algorithms,morella2023vertical, kaur2023comparative} that can anticipate the working conditions based on the data available from various modalities. However, these systems have been developed mainly for vertical hydroponic farms located in large spaces such as warehouses. Thus there is a dearth of robotic technologies developed specifically for CVFs due to the difficulty in representing the motion constraints on the robotic end-effector for different farming operations and also due to the limited space for operation. An excellent survey article~\cite{vasconez2019human} discusses human-robot interaction in agriculture and the potential use of cobots in conjunction with humans. The authors advocate for developing approaches for interaction between humans and cobots such that they are well suited to the needs of the task and the environment.  
Therefore, in order to overcome the issue of representing motion constraints associated with farming operations, like transplanting and harvesting, in constrained spaces, we propose using cobots alongside farmers in CVFs. The task-related motion constraints can be captured using \textit{kinesthetic demonstrations} provided to the cobot by a farmer.


Although there are a variety of approaches to learning from demonstration (see~\cite{ArgallCVB09, Billard2016, RavichandarPCB20} and the references therein), in CVF we are interested in methods in which the robot can even use a single example for planning.  An approach to generate motions from a single demonstration is the use of Dynamical Movement Primitives (DMP)~\cite{IjspeertNHPS13, HerschGCB08, pastor2009, Saveriano2019}, which is a dynamical systems-based approach.  While this is an elegant bio-inspired approach, DMPs do not consider end-effector motion constraints explicitly. Since satisfying end-effector motion constraints is critical to the tasks of transplantation and harvesting, we use a screw-geometric approach~\cite{mahalingam2023humanguided,laha2022}.

In~\cite{mahalingam2023humanguided}, we proposed an approach for extracting task-related constraints for complex manipulation tasks such as scooping and pouring using a \textit{single} kinesthetic demonstration by exploiting the screw geometric structure of motion. The extracted task-related constraints are stored as a sequence of constant screw motions (or one-parameter subgroups of $SE(3)$) that are coordinate invariant. This sequence of constant screw motions can then be used to generate motion plans for a different instance of the same task assuming the pose of the task-related objects is known. In this work, we use a similar approach for extracting and transferring the task-related constraints using kinesthetic demonstrations for the transplanting and harvesting operations inside a CVF. However in this work the pose of the task-related objects is not known beforehand.



%

\section{Mathematical Preliminaries}

\begin{figure*}[ht!]
    \centering
    \includegraphics[width=\textwidth]{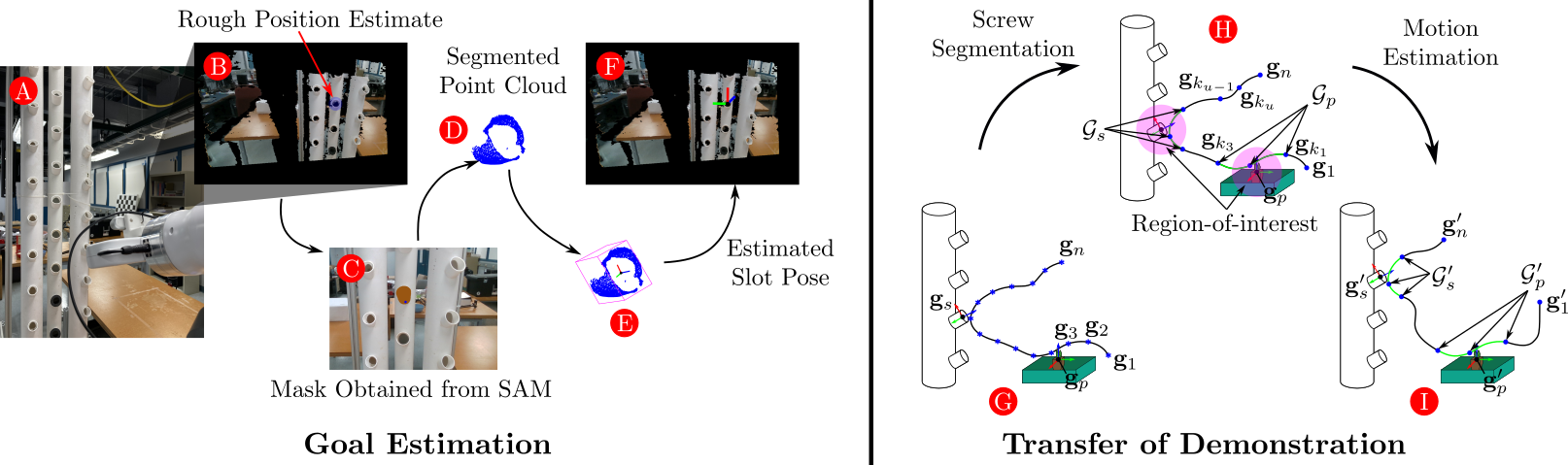}
    \caption{\textbf{Solution Approach Overview}: \textit{Left Image} - A) Eye-in-hand setup used to capture RGBD images B) Obtained RGBD image from sensor with the rough slot position estimate shown C) Position estimate of slot in the pixel space show as a blue dot and corresponding mask obtained from SAM shown in orange D) 3D points corresponding to slot segmented out of the RGBD image E) Bounding box fit to the 3D points corresponding to the slot for pose estimation F) Pose estimate of slot; \textit{Right Image} - Schematic sketch of motion estimation from demonstration for a transplanting task G) Recorded demonstration show with * used to respresent SE(3) poses to help reduce clutter H) The recorded demonstration is segmented into a sequence of constant screws for identifying the sequence of motion subgroup constraints on the end-effector motion relative to the task-related objects that lie within the region-of-interest of the plant sapling and planting slot I) Transferring the extracted constraints to a new planting slot and determining the final end-effector motion as a sequence of constant screws.}
    \label{fig:solution_approach}
\end{figure*}

In this section we provide a brief review of the background knowledge required for this work.
The joint space of the robot is the set of all possible joint configurations and is denoted by $\mathcal{J} \subset \mathbb{R}^n$ where $n$ is the number of degrees of freedom of the robot. The set of all rigid body configurations is $SE(3)$, the Special Euclidean Group of dimension $3$. An element of $SE(3)$ is also referred to as a \textit{pose}. Thus, the task space $\mathcal{T}$ of the robot, i.e., the set of end-effector poses, is a subset of $SE(3)$. The joint configuration of the robot, $\mathbf{\Theta} \in \mathcal{J}$ is a vector of length $n$. For every valid joint configuration $\mathbf{\Theta}$ the forward kinematics of the manipulator maps it to a unique end-effector pose, $\mathbf{g} \in \mathcal{T}$.

\noindent
{\bf Task Instance: }
Objects whose poses affect the generation of manipulation plans for performing a task are referred to as $\textit{task-relevant objects}$. A $\textit{task instance}$ is defined as the set of all task-relevant object poses and it is denoted by $\mathcal{O}$.

\noindent
{\bf Kinesthetic demonstration: } A user can provide a kinesthetic demonstration $\mathcal{D}$ by holding the robot end effector in zero gravity mode and performing the manipulation task. A particular demonstration $\mathcal{D}$ is associated with a task instance $\mathcal{O}$. The kinesthetic demonstration is recorded as a sequence of joint angles, i.e., a path in $\mathcal{J}$, which can be mapped to a path in the task space $\mathcal{T}$ using the forward kinematics map. Throughout this paper, whenever we mention a demonstration, we refer to the sequence of poses, $\mathcal{D} = \{\mathbf{g}_1, \mathbf{g}_2, \mathbf{g}_3, ..., \mathbf{g}_n\}$ in the task space $\mathcal{T}$ that the end-effector of the robot goes through while performing the task.

\noindent
\textbf{Screw Displacement}: Chasles-Mozzi theorem states that the general Euclidean displacement/motion of a rigid body from the origin $\boldsymbol{I}$ to $\boldsymbol{T} = (\boldsymbol{R},\boldsymbol{p}) \in SE(3)$
can be expressed as a rotation $\theta$ about a fixed axis $\mathcal{S}$, called the \textit{screw axis}, and a translation $d$ along that axis. 
Plücker coordinates can be used to represent the screw axis by $\boldsymbol{\omega}$ and $\boldsymbol{m}$, where $\boldsymbol{\omega} \in \mathbb{R}^3$ is a unit vector that represents the direction of the screw axis, $\boldsymbol{m} = \boldsymbol{r} \times \boldsymbol{\omega}$, and $\boldsymbol{r} \in \mathbb{R}^3$ is an arbitrary point on the screw axis. Thus, the screw parameters are defined as $\boldsymbol{\omega}, \boldsymbol{m}, h, \theta$, where $h$ is the pitch of the screw and $\theta$ is its magnitude. In general, for pure rotation and general screw motion, $h$ is finite, while for pure translation, $h = \infty$ with $\theta$ replaced by $d$. \textbf{A {\em constant screw motion} is a motion where the parameters $\boldsymbol{\omega}, \boldsymbol{m}$, and $h$ stays constant throughout the motion}.

\noindent
\textbf{Screw Linear Interpolation (ScLERP)}: To perform a one degree-of-freedom smooth screw motion (with a constant rotation and translation rate) between two object poses in $SE(3)$, Screw Linear interpolation (ScLERP) can be used. ScLERP generate a geodesic motion between two given poses in $SE(3)$.

\section{Problem Statement}

The goal of this paper is to study the feasibility of using robotic manipulators for automating tasks in CVF by using a single user provided kinesthetic demonstrations of such tasks. To evaluate this we have considered two tasks in this study: 1) Transplanting of saplings 2) Harvesting grown plants.

Using a single demonstration for each task (one demonstration for the transplanting task and one demonstration for the harvesting task), and sensor information from an RGBD camera, we want to transplant saplings into the growing tube and then harvest the grown plants from the growing tube. Note that any successful demonstration implicitly contains the constraints that characterize the tasks. We will now provide the problem statement for each task separately:

\noindent
\textbf{\underline{Transplanting Problem}: \textit{Given a single demonstration $\mathcal{D}$ of the transplanting task and an RGBD image $\mathcal{I}$ of the slot where the sapling is transplanted, taken before providing the demonstration, compute the motion plan, i.e., the sequence of joint configurations $\mathcal{M} = \{\Theta_1, \Theta_2, ..., \Theta_m$\} that would allow the robot to perform the transplanting task successfully when provided with the RGBD image $\mathcal{I}'$ of a different slot where the sapling has to be now planted.}}

For the harvesting task, when the plants are grown, the slots in the growing tubes and the base of the plants are occluded. In this case, we cannot use the image to estimate the pose where the plant should be grasped. However, as we will discuss in the next section, since during transplantation, we estimate a reference frame associated with a slot, we will use this to formulate our problem.

\noindent
\textbf{\underline{Harvesting Problem}: \textit{Given a single demonstration $\mathcal{D}$ of the harvesting task, and the reference frame of a different slot from which the plant has to be harvested, compute the motion plan $\mathcal{M} = \{\Theta_1, \Theta_2, ..., \Theta_m$\} that would allow the robot to perform the task successfully.}}

Note that in setting up the above problems, we assume that the tray containing the saplings to be transplanted and the collection tray for the harvested plants are located at a fixed pose within the robot workspace and known to the robot. Also, since the geometry of the growing tubes is similar (but not same), we use information such as the angle made by the axis of the cylindrical slot with the horizontal, the spacing between consecutive slots in the growing tube, and the distance between adjacent growing tubes on the panel to estimate the rough position of the slots within the workspace of the robot. 
These assumptions are reasonable since we are not modifying the environment in any way, just exploiting the existing structure of the environment in a CVF.

\section{Solution Approach}

\begin{figure*}[t!]
    \begin{subfigure}[b]{0.12\textwidth}
        \includegraphics[width=\textwidth]{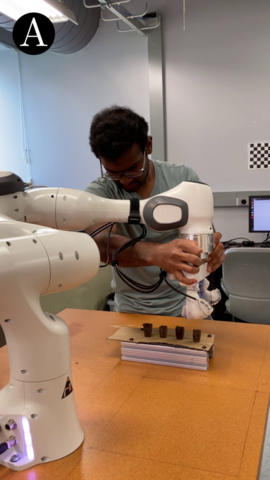}
    \end{subfigure}
    \begin{subfigure}[b]{0.12\textwidth}
        \includegraphics[width=\textwidth]{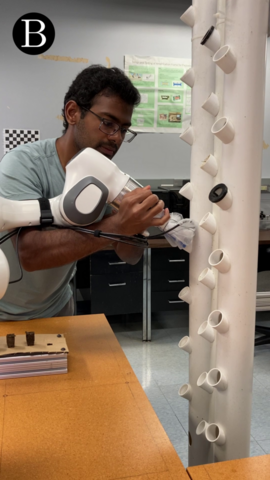}
    \end{subfigure}
    \begin{subfigure}[b]{0.12\textwidth}
        \includegraphics[width=\textwidth]{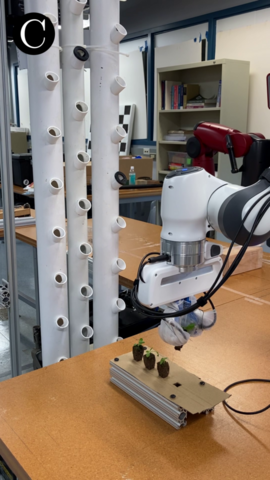}
    \end{subfigure}
    \begin{subfigure}[b]{0.12\textwidth}
        \includegraphics[width=\textwidth]{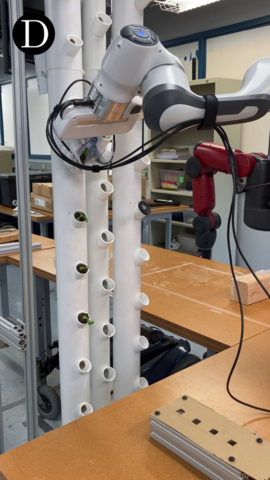}
    \end{subfigure}
    \begin{subfigure}[b]{0.12\textwidth}
        \includegraphics[width=\textwidth]{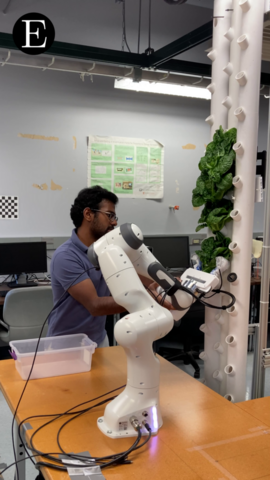}
    \end{subfigure}
    \begin{subfigure}[b]{0.12\textwidth}
        \includegraphics[width=\textwidth]{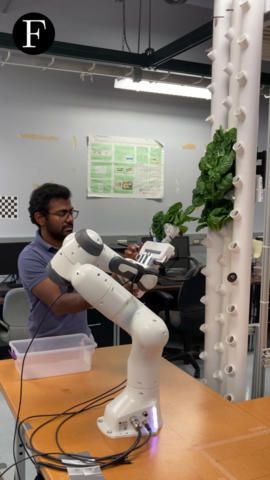}
    \end{subfigure}
    \begin{subfigure}[b]{0.12\textwidth}
        \includegraphics[width=\textwidth]{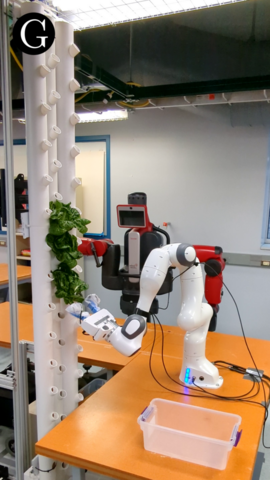}
    \end{subfigure}
    \begin{subfigure}[b]{0.12\textwidth}
        \includegraphics[width=\textwidth]{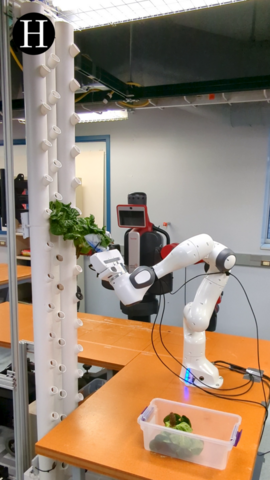}
    \end{subfigure}
    \caption{\textbf{Experimental Setup:} A, B: Demonstration for transplanting task, C, D: Execution from provided demonstration for transplanting task, E, F: Demonstration for harvesting task, G, H: Execution from provided demonstration for harvesting task}
    \label{fig:experimental_setup}
\end{figure*}

The key aspect of our solution approach is the use of a screw-geometry based representation of task constraints as proposed in \cite{mahalingam2023humanguided}. Furthermore, we also take advantage of the knowledge of the geometry of the growing tubes to estimate the pose of the slots in the growing tube. The screw-geometry based representation allows us to transfer the task constraints extracted from the demonstration relative to the pose of the new slot. These transferred screw constraints then allow us to use Screw Linear Interpolation (ScLERP) combined with Jacobian pseudo-inverse to compute a motion plan in the joint space \cite{sclerp_motion_planner}. The use of ScLERP ensures that the motion constraints present in the demonstration are transferred to the motion plan for the new task instance.
The three key steps in our solution approach are detailed below.

\noindent
\textbf{Screw Extraction: }
First, we extract the task constraints embedded in the demonstrated motion of the end-effector as a sequence of constant screws. The rationale behind the screw-geometry based representation of task constraints as stated in \cite{mahalingam2023humanguided} is two-fold. First, it can be inferred from Chasles' theorem than any path in $SE(3)$ can be approximated arbitrarily closely as a sequence of constant screw motions. This is analogous to the fact that any curve in $\mathbb{R}^3$ can be approximated arbitrarily closely by a sequence of straight line segments. Second, the screw representation is a coordinate-invariant representation (meaning that it does not depend on the choice of the coordinate frame at the end effector of the robot) and also maps the motion to a single parameter subgroup of $SE(3)$, which potentially allows better generalization properties. The extracted constant screw constraints allow us to express the demonstration $\mathcal{D} = \{\mathbf{g}_1, \mathbf{g}_2, \mathbf{g}_3, ..., \mathbf{g}_n\}$ as a sequence of constant screws $\mathcal{G} = \{\mathbf{g}_1, \mathbf{g}_{k_1}, \mathbf{g}_{k_2}, ..., \mathbf{g}_{k_u}, \mathbf{g}_n\}$. Here, every two consecutive poses in $\mathcal{G}$, $(\mathbf{g}_i, \mathbf{g}_{i+1}),~\text{where,}~i = 1, k_1, k_2, ..., k_u, n$ define a constant screw segment with the entire demonstration consisting of $u+1$ constant screw segments. The sequence of constant screws $\mathcal{G}$, is a sub-sequence of the demonstration $\mathcal{D}$ and $2 \leq k_i \leq (n-1)$ with $ k_i < k_{i+1} \forall~i = 1, 2, ..., (n-1)$.


\noindent
\textbf{Goal Estimation: }
To define the task instance associated with the demonstration or with the execution of a new instance, we need the poses of the task relevant objects. For the transplanting task, this consists of the pose of the planting slot, $\mathbf{g}_s$ in the growing tube and the pose of the pod containing the sapling, $\mathbf{g}_p$. Similarly, for the harvesting task, this consists of the the pose of the planting slot, $\mathbf{g}_s$ in the growing tube and the pose of the collection tray, $\mathbf{g}_t$. The pose of the pod containing the sapling, $\mathbf{g}_p$ and the pose of the collection tray, $\mathbf{g}_t$ are known because we assume that they are placed at a known location.
Since we do not have information about the exact pose of the slots, we need to estimate its pose for defining the task instance which would then allow us to express the task constraints relative to the pose of the slot.
We define the pose of a slot to be the pose of a bounding box that bounds the 3D points corresponding to that slot obtained from the RGBD image of the growing panel. We identify the 3D points corresponding to a slot by first segmenting the pixels corresponding to that slot in the RGB image and then de-projecting those pixels to $\mathbb{R}^3$ using the camera intrinsics. 
Given an RGB image, the Segment Anything Model (SAM) \cite{kirillov2023segmentanything} performs segmentation by generating the set of all possible masks on the image. A mask is a set of pixels corresponding to a specific region of the image. In order to extract the 3D points corresponding to a specific slot, we need to extract the corresponding mask from the set of all the masks generated by SAM. Since we are operating in a structured environment with known camera extrinsics, a rough estimate of the slot position in $\mathbb{R}^3$ is projected back on to the pixel space i.e., the RGB image. This location of a particular slot in the pixel space is used to extract the mask corresponding to the slot. A sample output of this process is shown in Figure \ref{fig:solution_approach}-C where the position estimate of slot in the pixel space is shown in blue and the corresponding mask is shown in orange.


The pose of the planting slot, $\mathbf{g}_s$ obtained through this process allows us to define the task instance for the transplanting task, $\mathcal{O}_t = \{\mathbf{g}_p, \mathbf{g}_s\}$.
While it is possible to obtain the RGBD image of the growing panel before performing the transplanting task, it is however difficult to obtain one before performing the harvesting task as the foliage of the grown plants occlude the growing panel. But, due to the fact that a plant can be harvested only after it is transplanted, the pose estimate of the slot obtained before the transplanting task can be used to define the task instance of harvesting, $\mathcal{O}_h = \{\mathbf{g}_s, \mathbf{g}_t\}$ from the same slot. In this manner, for every slot in which we transplant a sapling, we also define a task instance of harvesting from that corresponding slot to be used for performing harvesting in the future.

\noindent
\textbf{Transfer of Demonstration: }
Recall that we have a demonstration $\mathcal{D} = \{\mathbf{g}_1, \mathbf{g}_2, \mathbf{g}_3, ..., \mathbf{g}_n\}$ of the transplanting task and suppose that we have identified the task instance $\mathcal{O}_t = \{\mathbf{g}_p, \mathbf{g}_s\}$ associated with this demonstration. For transferring the demonstration, we need to identify the task relevant constraints that need to be followed to successfully execute this task as a sequence of motion subgroup constraints on the end-effector motion relative to the task-related objects. Based on the heuristic defined in \cite{mahalingam2023humanguided}, we follow the definition of task-relevant constraints as, \textbf{\textit{the sequence of motion subgroup constraints obtained from the provided demonstration that need to be enforced relative to the task-related object within a region-of-interest surrounding them}}. So we construct the region-of-interest as a sphere surrounding each task-related object, which are the sapling and the planting slot, to identify the task-relevant constraints that lie within this sphere. 

Using the sequence of constant screws, $\mathcal{G} = \{\mathbf{g}_1, \mathbf{g}_{k_1}, \mathbf{g}_{k_2}, ..., \mathbf{g}_{k_u}, \mathbf{g}_n\}$, extracted from the demonstration $\mathcal{D}$, we identify the constant screws that lie inside the region-of-interest of the sapling as $\{\mathbf{g}_{k_{p1}}, \mathbf{g}_{k_{p2}}, ..., \mathbf{g}_{k_{pl}}\}$ and the ones lying inside the region-of-interest of the planting slot as $\{\mathbf{g}_{k_{s1}}, \mathbf{g}_{k_{s2}}, ..., \mathbf{g}_{k_{sl}}\}$. Both the sequences are disjoint sub-sequences of $\mathcal{G}$ with, $k_1 \leq k_{p1} < k_{pl} < k_{s1} < k_{sl} \leq k_u$. We express them relative to the local frame of reference as $\mathcal{G}_p = \{\mathbf{g}_p^{-1}\mathbf{g}_{k_{p1}}, \mathbf{g}_p^{-1}\mathbf{g}_{k_{p2}}, ..., \mathbf{g}_p^{-1}\mathbf{g}_{k_{pl}}\}$ and $\mathcal{G}_s = \{\mathbf{g}_s^{-1}\mathbf{g}_{k_{s1}}, \mathbf{g}_s^{-1}\mathbf{g}_{k_{s2}}, ..., \mathbf{g}_s^{-1}\mathbf{g}_{k_{sl}}\}$. Given a new task instance $(\mathbf{g}_p', \mathbf{g}_s')$ of transplanting into a different slot, we can now determine the sequence of motion subgroup constraints on the end-effector motion as $\mathcal{G}_p' = \{\mathbf{g}_p'\mathbf{g}_p^{-1}\mathbf{g}_{k_{p1}}, \mathbf{g}_p'\mathbf{g}_p^{-1}\mathbf{g}_{k_{p2}}, ..., \mathbf{g}_p'\mathbf{g}_p^{-1}\mathbf{g}_{k_{pl}}\}$ and $\mathcal{G}_s' = \{\mathbf{g}_s'\mathbf{g}_s^{-1}\mathbf{g}_{k_{s1}}, \mathbf{g}_s'\mathbf{g}_s^{-1}\mathbf{g}_{k_{s2}}, ..., \mathbf{g}_s'\mathbf{g}_s^{-1}\mathbf{g}_{k_{sl}}\}$. We can now construct the sequence of constant screws that the end-effector needs to follow to execute the task as $\mathcal{G}' = \{\mathcal{G}_p', \mathcal{G}_s'\}$. Figure \ref{fig:solution_approach}-G,H,I show a schematic sketch of motion estimation from demonstration for a new task instance of the transplanting task.
A similar approach can be followed for the transfer of demonstration of the harvesting task.

\section{Experimental Results}

\begin{figure}[h!]
  \centering
  \includegraphics[width=\linewidth]{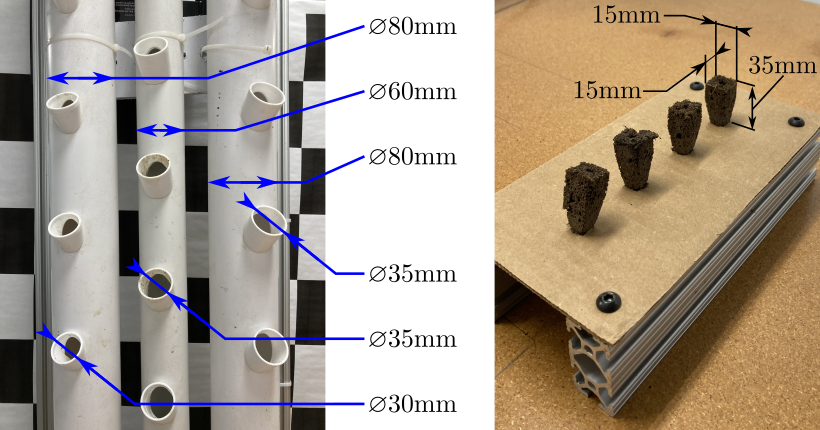}
  \caption{\textit{Left Image} - View of growing panel from front showing pipe and slot dimensions; \textit{Right Image} - View of tray containing sapling pods showing the dimensions of the sapling pods}
 \label{fig:pipe_and_pod_dimensions} 
\end{figure}

To evaluate our approach we conducted experimental trials for both tasks. The experiments were carried out in a laboratory with the Franka Emika Panda manipulator fixed on a table and a CVF setup consisting of a grow panel with three vertical growing tubes in front of the robot (see Figure \ref{fig:solution_approach}, \ref{fig:experimental_setup}). The robot was equipped with a standard parallel jaw gripper to grasp the objects. Each growing tube had multiple plantings slots into (from) which the sapling (grown plant) should be planted (harvested). We used only the planting slots in the growing tubes that were within the workspace of the manipulator as the robot base was fixed to the table and the growing tubes are stationary. We use an Intel Realsense D415 camera in an eye-in-hand configuration for obtaining sensor data in form of RGBD images. The growing panel arrangement used three tubes, each of a different specification and two types of slot specification. The different tube and slot specifications are show in Figure \ref{fig:pipe_and_pod_dimensions}. The reason for the variability in the specifications was to evaluate the performance of the robot against different specifications used in CVF. Also the dimensions of the pods containing the saplings are show in Figure \ref{fig:pipe_and_pod_dimensions}.

We provided multiple demonstrations for each task, and using each demonstration we conducted multiple trials. Figure \ref{fig:experimental_setup} shows the experimental setup. Figure \ref{fig:experimental_setup}-A, B shows how the demonstrations were provided for the transplanting task. Figure \ref{fig:experimental_setup}-C, D shows the execution of the transplanting task using the provided demonstration. Figure \ref{fig:experimental_setup}-E, F shows how the demonstrations were provided for the harvesting task. Figure \ref{fig:experimental_setup}-G, H shows the execution of the harvesting task using the provided demonstration. The results of the conducted experiments are summarised in Table \ref{exp_trials}. We observed an overall success rate of 83.8\% when using our approach. Among the trials conducted for both tasks, for each demonstration we included trials on a growing tube of a specification that was different from the one using which the demonstration was provided. The robot performed the tasks successfully even under such variations. This shows the robustness of our approach.

\noindent
\textbf{Transplanting Task: }
The main takeaway from the results of the transplanting task is that using the screw-geometry based representation of task constraints combined with the estimation of the goal using sensor data it is possible to capture the task constraints and use it for generating the motion plan for a new task instance even when there are tight tolerances such as in the case of inserting the pod containing the sapling into the slot in the growing tube. Even though point cloud obtained from the RGBD sensor is noisy and has some occlusions (See Figure \ref{fig:solution_approach}), we were able to insert the sapling pod which has a square cross-section of 15 mm at its widest part into the planting slots of 30 mm and 35 mm diameter while also satisfying the task constraints. We observed two failures for this task. Both failures were due to the combined effects of gripper geometry and error in the pose estimate of the planting slots.
In one case, the pod was not fully inserted in the slot, and in the other case the fingertip of the robot gripper hit the planting slot and was unable to insert the pod into the slot. We believe that changing the gripper geometry by making it thinner and longer can make the insertion process more robust and increase the success rate.

\noindent
\textbf{Harvesting Task: }
For the harvesting task since the foliage of the fully grown plants occlude the growing tubes completely, it is hard to use RGBD sensor information to identify the location of the slots that would allow us to grasp the plant at the base of the main stem. However, as stated before, we can use the estimate of the slot obtained during transplanting to harvest the fully grown plant. So using the pose estimates of the planting slots from the transplanting task, we perform the harvesting task. For this task we observed only one type of failure. All failures were due to leaves of the neighbouring plant getting caught in between the gripper tips during harvesting of a grown plant. This causes the neighbouring plant to also be harvested but slip from the gripper after being pulled out of the planting slot and eventually fall down. This issue could also be potentially resolved by better gripper design choices.

\begin{table}[ht!]
\centering
\renewcommand{\arraystretch}{1.25}
\begin{tabular}{|c|c|c|c|} 
    \hline
    \textbf{Task} & \multirow{1}{*}{\textbf{Demo No.}} & 
    \textbf{No. of Trials} & 
    \textbf{Successful}\\
    \hline
    \hline
    \multirow{3}{*}{Transplanting} & Demo \#1 & 5 & 4\\
    \cline{2-4}
    & Demo \#2 & 5 & 4\\
    \cline{2-4}
    & Demo \#3 & 5 & 5\\
    \hline
    \multirow{4}{*}{Harvesting} & Demo \#1 & 4 & 3\\
    \cline{2-4}
    & Demo \#2 & 4 & 3\\
    \cline{2-4}
    & Demo \#3 & 4 & 4\\
    \cline{2-4}
    & Demo \#4 & 4 & 3\\
    \hline
\end{tabular}
\caption{Experimental Trials}
\label{exp_trials}
\end{table}

\section{Conclusion}
In this paper, we present a novel method for performing two key tasks that arise in containerized vertical farming, namely, sapling transplantation and harvesting, using a cobot arm. Both tasks are characterized by the presence of end-effector motion constraints, which are difficult to specify and have to be estimated based on image data. This makes it challenging to apply constrained motion planning algorithms and learning from demonstration algorithms that do not try to extract the constraints present in the demonstration. We show that representing the constraints present in a demonstration as motion subgroups in $SE(3)$ is a viable way to plan for these tasks. We presented a method to use the extracted motion subgroup constraints to generate the motion subgroup constraints for new task instances based on RGBD images of the scene and geometric knowledge of the growing tubes. As motion subgroup constraints can be satisfied using ScLERP-based planners~\cite{sclerp_motion_planner}, we can ensure that the generated motion plans satisfy the constraints characterizing the tasks. Our experimental results, conducted with a Franka Emika Panda robot with a wrist-mounted RGBD camera and a parallel jaw gripper, showed a success rate of about $83.8\%$. The failures were mainly due to the geometry of the gripper (i.e., the thickness and length of the fingers). We believe that our results can be further improved with customized grippers with different finger length and thickness, and optimizing the gripper dimensions for these tasks is a future avenue of research that we will pursue.       

\nocite{Murray1994}

\bibliographystyle{IEEEtran}
\bibliography{vertical_farming, references}

\begin{thebibliography}{10}
\providecommand{\url}[1]{#1}
\csname url@rmstyle\endcsname
\providecommand{\newblock}{\relax}
\providecommand{\bibinfo}[2]{#2}
\providecommand\BIBentrySTDinterwordspacing{\spaceskip=0pt\relax}
\providecommand\BIBentryALTinterwordstretchfactor{4}
\providecommand\BIBentryALTinterwordspacing{\spaceskip=\fontdimen2\font plus
\BIBentryALTinterwordstretchfactor\fontdimen3\font minus
  \fontdimen4\font\relax}
\providecommand\BIBforeignlanguage[2]{{%
\expandafter\ifx\csname l@#1\endcsname\relax
\typeout{** WARNING: IEEEtran.bst: No hyphenation pattern has been}%
\typeout{** loaded for the language `#1'. Using the pattern for}%
\typeout{** the default language instead.}%
\else
\language=\csname l@#1\endcsname
\fi
#2}}

\bibitem{birkby2016vertical}
J.~Birkby, ``Vertical farming,'' \emph{ATTRA sustainable agriculture}, vol.~2,
  pp. 1--12, 2016.

\bibitem{marchant2017robotic}
W.~Marchant and S.~Tosunoglu, ``Robotic implementation to automate a vertical
  farm system,'' in \emph{Proceedings of the 30th Florida Conference on Recent
  Advances in Robotics}, 2017, pp. 11--12.

\bibitem{plenty_main}
{{Plenty Farms}}, \url{https://www.plenty.ag/mission/}.

\bibitem{plenty_tigris_1}
{{Plenty Farms Tigris Facility}},
  \url{https://www.youtube.com/watch?v=fb4xcFw2VMg}.

\bibitem{bowery_main}
{{Bowery Farms}}, \url{https://bowery.co/about-us/}.

\bibitem{bowery_bloomberg}
{{Bloomberg Originals: The High-Tech Vertical Farmer}},
  \url{https://www.youtube.com/watch?v=AGcYApKfHuY}.

\bibitem{bowery_tech_and_innovation}
{{Tech and Innovation at Bowery Farms}},
  \url{https://www.youtube.com/watch?v=VVi6-FAtMcU}.

\bibitem{kirillov2023segmentanything}
A.~Kirillov, E.~Mintun, N.~Ravi, H.~Mao, C.~Rolland, L.~Gustafson, T.~Xiao,
  S.~Whitehead, A.~C. Berg, W.-Y. Lo, \emph{et~al.}, ``Segment anything,''
  \emph{arXiv preprint arXiv:2304.02643}, 2023.

\bibitem{mahalingam2023humanguided}
D.~Mahalingam and N.~Chakraborty, ``Human-guided planning for complex
  manipulation tasks using the screw geometry of motion,'' in \emph{2023 IEEE
  International Conference on Robotics and Automation (ICRA)}, May 2023, pp.
  7851--7857.

\bibitem{sclerp_motion_planner}
A.~Sarker, A.~Sinha, and N.~Chakraborty, ``On screw linear interpolation for
  point-to-point path planning,'' in \emph{2020 IEEE/RSJ International
  Conference on Intelligent Robots and Systems (IROS)}, 2020, pp. 9480--9487.

\bibitem{bac2014harvesting}
C.~W. Bac, E.~J. Van~Henten, J.~Hemming, and Y.~Edan, ``Harvesting robots for
  high-value crops: State-of-the-art review and challenges ahead,''
  \emph{Journal of Field Robotics}, vol.~31, no.~6, pp. 888--911, 2014.

\bibitem{roldan2018robots}
J.~J. Rold{\'a}n, J.~del Cerro, D.~Garz{\'o}n-Ramos, P.~Garcia-Aunon,
  M.~Garz{\'o}n, J.~De~Le{\'o}n, and A.~Barrientos, ``Robots in agriculture:
  State of art and practical experiences,'' \emph{Service robots}, pp. 67--90,
  2018.

\bibitem{vasconez2019human}
J.~P. Vasconez, G.~A. Kantor, and F.~A.~A. Cheein, ``Human--robot interaction
  in agriculture: A survey and current challenges,'' \emph{Biosystems
  engineering}, vol. 179, pp. 35--48, 2019.

\bibitem{van2017multisensor}
M.~van Dael, P.~Verboven, J.~Dhaene, L.~Van~Hoorebeke, J.~Sijbers, and
  B.~Nicolai, ``Multisensor x-ray inspection of internal defects in
  horticultural products,'' \emph{Postharvest biology and technology}, vol.
  128, pp. 33--43, 2017.

\bibitem{corollaro2014combined}
M.~L. Corollaro, E.~Aprea, I.~Endrizzi, E.~Betta, M.~L. Dematt{\`e},
  M.~Charles, M.~Bergamaschi, F.~Costa, F.~Biasioli, L.~C. Grappadelli,
  \emph{et~al.}, ``A combined sensory-instrumental tool for apple quality
  evaluation,'' \emph{Postharvest Biology and Technology}, vol.~96, pp.
  135--144, 2014.

\bibitem{pace2011relationship}
B.~Pace, M.~Cefola, F.~Renna, and G.~Attolico, ``Relationship between visual
  appearance and browning as evaluated by image analysis and chemical traits in
  fresh-cut nectarines,'' \emph{Postharvest Biology and Technology}, vol.~61,
  no. 2-3, pp. 178--183, 2011.

\bibitem{munera2017astringency}
S.~Munera, C.~Besada, J.~Blasco, S.~Cubero, A.~Salvador, P.~Talens, and
  N.~Aleixos, ``Astringency assessment of persimmon by hyperspectral imaging,''
  \emph{Postharvest Biology and Technology}, vol. 125, pp. 35--41, 2017.

\bibitem{mishraIoT2022}
N.~N. Misra, Y.~Dixit, A.~Al-Mallahi, M.~S. Bhullar, R.~Upadhyay, and
  A.~Martynenko, ``Iot, big data, and artificial intelligence in agriculture
  and food industry,'' \emph{IEEE Internet of Things Journal}, vol.~9, no.~9,
  pp. 6305--6324, 2022.

\bibitem{ukwuoma2022recent}
C.~C. Ukwuoma, Q.~Zhiguang, M.~B. Bin~Heyat, L.~Ali, Z.~Almaspoor, and H.~N.
  Monday, ``Recent advancements in fruit detection and classification using
  deep learning techniques,'' \emph{Mathematical Problems in Engineering}, vol.
  2022, pp. 1--29, 2022.

\bibitem{tang2020recognition}
Y.~Tang, M.~Chen, C.~Wang, L.~Luo, J.~Li, G.~Lian, and X.~Zou, ``Recognition
  and localization methods for vision-based fruit picking robots: A review,''
  \emph{Frontiers in Plant Science}, vol.~11, p. 510, 2020.

\bibitem{nuske2011visual}
S.~Nuske, S.~Achar, K.~Gupta, S.~Narasimhan, and S.~Singh, ``Visual yield
  estimation in vineyards: experiments with different varietals and calibration
  procedures,'' \emph{Robotics Institute, Carnegie Mellon University Technical
  Report}, 2011.

\bibitem{nuske2011yield}
S.~Nuske, S.~Achar, T.~Bates, S.~Narasimhan, and S.~Singh, ``Yield estimation
  in vineyards by visual grape detection,'' in \emph{2011 IEEE/RSJ
  International Conference on Intelligent Robots and Systems}.\hskip 1em plus
  0.5em minus 0.4em\relax IEEE, 2011, pp. 2352--2358.

\bibitem{de2011design}
Z.~De-An, L.~Jidong, J.~Wei, Z.~Ying, and C.~Yu, ``Design and control of an
  apple harvesting robot,'' \emph{Biosystems engineering}, vol. 110, no.~2, pp.
  112--122, 2011.

\bibitem{bagagiolo2022greenhouse}
G.~Bagagiolo, G.~Matranga, E.~Cavallo, and N.~Pampuro, ``Greenhouse robots:
  Ultimate solutions to improve automation in protected cropping systems—a
  review,'' \emph{Sustainability}, vol.~14, no.~11, p. 6436, 2022.

\bibitem{van2013robotics}
E.~Van~Henten, C.~Bac, J.~Hemming, and Y.~Edan, ``Robotics in protected
  cultivation,'' \emph{IFAC Proceedings Volumes}, vol.~46, no.~18, pp.
  170--177, 2013.

\bibitem{sa2017peduncle}
I.~Sa, C.~Lehnert, A.~English, C.~McCool, F.~Dayoub, B.~Upcroft, and T.~Perez,
  ``Peduncle detection of sweet pepper for autonomous crop
  harvesting—combined color and 3-d information,'' \emph{IEEE Robotics and
  Automation Letters}, vol.~2, no.~2, pp. 765--772, 2017.

\bibitem{vitzrabin2016changing}
E.~Vitzrabin and Y.~Edan, ``Changing task objectives for improved sweet pepper
  detection for robotic harvesting,'' \emph{IEEE Robotics and Automation
  Letters}, vol.~1, no.~1, pp. 578--584, 2016.

\bibitem{aastrand2002agricultural}
B.~{\AA}strand and A.-J. Baerveldt, ``An agricultural mobile robot with
  vision-based perception for mechanical weed control,'' \emph{Autonomous
  robots}, vol.~13, pp. 21--35, 2002.

\bibitem{slaughter2008autonomous}
D.~C. Slaughter, D.~Giles, and D.~Downey, ``Autonomous robotic weed control
  systems: A review,'' \emph{Computers and electronics in agriculture},
  vol.~61, no.~1, pp. 63--78, 2008.

\bibitem{heravi2018development}
A.~Heravi, D.~Ahmad, I.~A. Hameed, R.~R. Shamshiri, S.~K. Balasundram, and
  M.~Yamin, ``Development of a field robot platform for mechanical weed control
  in greenhouse cultivation of cucumber,'' \emph{Agricultural robots:
  fundamentals and applications}, pp. 11--29, 2018.

\bibitem{warrier2021design}
K.~Warrier, M.~Rajendiran, S.~K. Kannan, and R.~Ranjith~Pillai, ``Design and
  development of automated vertical farming setup,'' in \emph{Handbook of Smart
  Materials, Technologies, and Devices: Applications of Industry 4.0}.\hskip
  1em plus 0.5em minus 0.4em\relax Springer, 2021, pp. 1--36.

\bibitem{lauguico2019implementation}
S.~C. Lauguico, R.~S. Concepcion, D.~D. Macasaet, J.~D. Alejandrino, A.~A.
  Bandala, and E.~P. Dadios, ``Implementation of inverse kinematics for
  crop-harvesting robotic arm in vertical farming,'' in \emph{2019 IEEE
  International Conference on Cybernetics and Intelligent Systems (CIS) and
  IEEE Conference on Robotics, Automation and Mechatronics (RAM)}.\hskip 1em
  plus 0.5em minus 0.4em\relax IEEE, 2019, pp. 298--303.

\bibitem{chitre2023optimal}
N.~Chitre, A.~Dogra, and E.~Singla, ``Optimal synthesis of reconfigurable
  manipulators for robotic assistance in vertical farming,'' \emph{Robotica},
  pp. 1--15, 2023.

\bibitem{sreedevi2020digital}
T.~Sreedevi and M.~S. Kumar, ``Digital twin in smart farming: A categorical
  literature review and exploring possibilities in hydroponics,'' \emph{2020
  Advanced Computing and Communication Technologies for High Performance
  Applications (ACCTHPA)}, pp. 120--124, 2020.

\bibitem{samaranayake2022autonomous}
S.~Samaranayake, S.~Krishmal, P.~Cooray, T.~Senatilaka, S.~Rajapaksha, and
  W.~S. Nuwanthika, ``Autonomous hydroponic environment with live remote
  consulting system for strawberry farming,'' in \emph{2022 4th International
  Conference on Advancements in Computing (ICAC)}.\hskip 1em plus 0.5em minus
  0.4em\relax IEEE, 2022, pp. 54--59.

\bibitem{gubanov2022algorithms}
B.~Gubanov, V.~Lebedeva, I.~Lebedev, and M.~Astapova, ``Algorithms and software
  for evaluation of plant height in vertical farm using uavs,'' in
  \emph{Agriculture Digitalization and Organic Production: Proceedings of the
  Second International Conference, ADOP 2022, St. Petersburg, Russia, June
  06--08, 2022}.\hskip 1em plus 0.5em minus 0.4em\relax Springer, 2022, pp.
  351--362.

\bibitem{morella2023vertical}
P.~Morella, M.~P. Lamb{\'a}n, J.~Royo, and J.~C. S{\'a}nchez, ``Vertical
  farming monitoring: How does it work and how much does it cost?''
  \emph{Sensors}, vol.~23, no.~7, p. 3502, 2023.

\bibitem{kaur2023comparative}
G.~Kaur, P.~Upadhyaya, and P.~Chawla, ``Comparative analysis of iot-based
  controlled environment and uncontrolled environment plant growth monitoring
  system for hydroponic indoor vertical farm,'' \emph{Environmental Research},
  vol. 222, p. 115313, 2023.

\bibitem{ArgallCVB09}
B.~D. Argall, S.~Chernova, M.~Veloso, and B.~Browning, ``A survey of robot
  learning from demonstration,'' \emph{Robotics and Autonomous Systems},
  vol.~57, pp. 469 -- 483, 2009.

\bibitem{Billard2016}
\BIBentryALTinterwordspacing
A.~G. Billard, S.~Calinon, and R.~Dillmann, \emph{Learning from Humans}.\hskip
  1em plus 0.5em minus 0.4em\relax Cham: Springer International Publishing,
  2016, pp. 1995--2014. [Online]. Available:
  \url{https://doi.org/10.1007/978-3-319-32552-1_74}
\BIBentrySTDinterwordspacing

\bibitem{RavichandarPCB20}
H.~Ravichandar, A.~S. Polydoros, S.~Chernova, and A.~Billard, ``Recent advances
  in robot learning from demonstration,'' \emph{Annual Review of Control,
  Robotics, and Autonomous Systems}, vol.~3, no.~1, pp. 297--330, 2020.

\bibitem{IjspeertNHPS13}
\BIBentryALTinterwordspacing
A.~J. Ijspeert, J.~Nakanishi, H.~Hoffmann, P.~Pastor, and S.~Schaal,
  ``Dynamical movement primitives: Learning attractor models for motor
  behaviors,'' \emph{Neural Comput.}, vol.~25, no.~2, pp. 328--373, Feb. 2013.
  [Online]. Available: \url{http://dx.doi.org/10.1162/NECO_a_00393}
\BIBentrySTDinterwordspacing

\bibitem{HerschGCB08}
M.~Hersch, F.~Guenter, S.~Calinon, and A.~Billard, ``Dynamical system
  modulation for robot learning via kinesthetic demonstrations,'' \emph{IEEE
  Transactions on Robotics}, vol.~24, no.~6, pp. 1463--1467, Dec 2008.

\bibitem{pastor2009}
P.~Pastor, H.~Hoffmann, T.~Asfour, and S.~Schaal, ``Learning and generalization
  of motor skills by learning from demonstration,'' in \emph{2009 IEEE
  International Conference on Robotics and Automation}, May 2009, pp. 763--768.

\bibitem{Saveriano2019}
M.~Saveriano, F.~Franzel, and D.~Lee, ``Merging position and orientation motion
  primitives,'' in \emph{2019 International Conference on Robotics and
  Automation (ICRA)}, May 2019, pp. 7041--7047.

\bibitem{laha2022}
R.~Laha, R.~Sun, W.~Wu, D.~Mahalingam, N.~Chakraborty, L.~F. Figueredo, and
  S.~Haddadin, ``Coordinate invariant user-guided constrained path planning
  with reactive rapidly expanding plane-oriented escaping trees,'' in
  \emph{2022 International Conference on Robotics and Automation (ICRA)}, May
  2022, pp. 977--984.

\bibitem{Murray1994}
R.~M. Murray, Z.~Li, and S.~S. Sastry, \emph{A mathematical introduction to
  robotic manipulation}.\hskip 1em plus 0.5em minus 0.4em\relax CRC press,
  1994.

\end{thebibliography}

\end{document}